\documentclass[]{cas-dc}

\usepackage[numbers]{natbib}

\def\tsc#1{\csdef{#1}{\textsc{\lowercase{#1}}\xspace}}
\tsc{WGM}
\tsc{QE}
\tsc{EP}
\tsc{PMS}
\tsc{BEC}
\tsc{DE}

\begin{document}
\let\WriteBookmarks\relax
\def\floatpagepagefraction{1}
\def\textpagefraction{.001}

\shorttitle{GUDN}

\shortauthors{Qing Wang et~al.}

\title [mode = title]{GUDN: A novel guide network with label reinforcement strategy for extreme multi-label text classification}

\tnotetext[1]{This document is the results of the research
   project funded by the National Science Foundation.}

\tnotetext[2]{The second title footnote which is a longer text matter
   to fill through the whole text width and overflow into
   another line in the footnotes area of the first page.}

%

\author[1]{Qing Wang}[type=editor,
                        auid=000,bioid=1,
                        orcid=]

\ead{wq2481@zjnu.edu.cn}

\credit{Conceptualization of this study, Methodology, Software}

\address{
    Zhejiang Normal University,
    No. 688 Yingbin Avenue,
    Jinhua City,
    321004,
    China}

\address{
    Stanford University,
    450 Serra Mall, Stanford,
    CA 94305,
    California,
    USA}

\author[1,2]{Jia Zhu}[style=chinese]

\cormark[1]

\ead{jiazhu@zjnu.edu.cn and zhujia@stanford.edu}
\author[1]{Hongji Shu}
\ead{shj451148969@zjnu.edu.cn}

\author[1]{Kwame Omono Asamoah}

\ead{koasamoah2014@gmail.com}

\author[1]{Jianyang Shi}
\ead{shijianyang@zjnu.edu.cn}

\author[1]{Cong Zhou}
\ead{zhoucong@zjnu.edu.cn}

\cortext[cor1]{Corresponding author}
\cortext[cor2]{Principal corresponding author}

\fntext[fn1]{This is the first author footnote. but is common to third
  author as well.}
\fntext[fn2]{Another author footnote, this is a very long footnote and
  it should be a really long footnote. But this footnote is not yet
  sufficiently long enough to make two lines of footnote text.}

\nonumnote{This note has no numbers. In this work we demonstrate $a_b$
  the formation Y\_1 of a new type of polariton on the interface
  between a cuprous oxide slab and a polystyrene micro-sphere placed
  on the slab.
  }

\begin{abstract}
In natural language processing, extreme multi-label text classification is an emerging but essential task. The problem of extreme multi-label text classification (XMTC) is to recall some of the most relevant labels for a text from an extremely large label set. Large-scale pre-trained models have brought a new trend to this problem. Though the large-scale pre-trained models have made significant achievements on this problem, the valuable fine-tuned methods have yet to be studied. Though label semantics have been introduced in XMTC, the vast semantic gap between texts and labels has yet to gain enough attention.
This paper builds a new guide network (GUDN) to help fine-tune the pre-trained model to instruct classification later.
Furthermore, GUDN uses raw label semantics combined with a helpful label reinforcement strategy to effectively explore the latent space between texts and labels, narrowing the semantic gap, which can further improve predicted accuracy. Experimental results demonstrate that GUDN outperforms state-of-the-art methods on Eurlex-4k and has competitive results on other popular datasets. In an additional experiment, we investigated the input lengths' influence on the Transformer-based model's accuracy. Our source code is released at https://t.hk.uy/aFSH.
\end{abstract}


\begin{highlights}
\item Research highlights item 1
\item Research highlights item 2
\item Research highlights item 3
\end{highlights}


\maketitle

\section{Introduction} 
\begin{figure}[!htb]
\centering
\includegraphics[scale = 0.48]{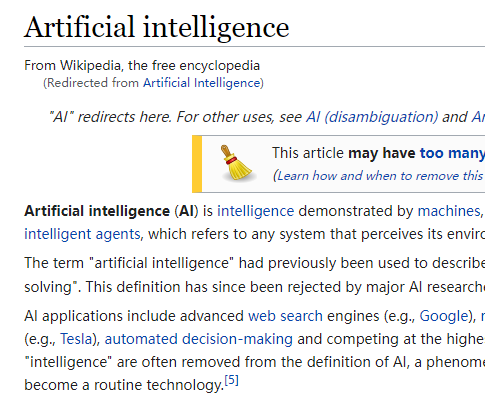}
\caption{An example of XMTC is extracted from https://en.wikipedia.org/wiki/Artificial\_intelligence.}
\label{ai}
\end{figure}

The goal of extreme multi-label text classification (XMTC) is to find the most relevant labels for a text among a large number of labels. It is a ubiquitous problem, from e-commerce platforms to term search sites. Figure \ref{ai} depicts an article from Wikipedia introducing artificial intelligence. We can find many artificial-intelligence-related labels in this article, such as ``Google'', ``automated decision-making'', ``machines'', and even other notions of interdisciplinary fields (usually on Wikipedia, the number of these terms or labels can reach hundreds of thousands). These labels, when clicked on, give access to the related articles. The number of labels keeps increasing; hence, the XMTC method is needed to match the articles and the labels. The core challenge with the XMTC is how to match these labels to the correct texts quickly and accurately.

It has been emphasized that extreme multi-label text classification differs from multi-class or multi-label text classification. The labels in XMTC tasks can reach hundreds of thousands or even more, impairing predictive accuracy and increasing computation time. The ample but sparse label space has made the ``long-tail'' distribution apparent leading to poor accuracy in some samples. In this situation, the length of texts is also long, requiring a lot of memory and time for training. The XMTC has attracted much research interest in the last decade due to its wide downstream applications, like advertising, user profiles, and web search.

Many traditional machine learning methods, including FastXML \cite{2014FastXML}, Bloom Filters \cite{cisse2013robust}, and Dismec \cite{babbar2016dismec}, have been proposed to solve the XMTC problem, with relatively impressive results in some ways. However, many of these methods often rely upon some specific setting with low efficiency. For instance, Bloom Filters is only feasible when the label matrix is low rank. Moreover, they often use sparse features like the TF-IDF and Bag of Words (BOW) as the input, needing more semantic information, which is hard to optimize. They need to break through the limitations of traditional methods to achieve high prediction accuracy.

Deep learning methods have started to flourish in XMTC tasks in recent years. One example of a representative method is C2AE \cite{2017Learning}, which uses a sparse linear network to explore the latent space between texts and labels. RankAE \cite{2019Ranking} further improved C2AE and spread it to extreme multi-label text classification. Some recent works, such as DECAF \cite{2021arXiv210800368M}, indicate that label metadata such as label structure and description are helpful for XMTC, so the effective extraction of descriptive label semantic features can provide an immediate performance boost. Nevertheless, C2AE and RankAE ignore the label metadata, barely using the multi-hot label representation vector to explore the latent space between text and label. Additionally, more than just using sparse linear layers to find latent space is required.

DXML \cite{2018Deep} used the label structure to build a graph and then revealed the potential relationship between texts and labels. LAHA \cite{2019Label} further introduces the attention mechanism for label embedding. Even though DXML and LAHA look at the metadata of the labels, the graph structure hides the original feature of the labels. Moreover, DXML and LAHA did not note the semantic gap between text and label. From the perspective of natural language, there must be some connection between text and label semantics. However, these connections have yet to be precisely exploited, failing to fix this issue. Furthermore, they need to pay more attention to the substantial semantic gap between texts and labels.

Some recent work adopts the large-scale pre-trained model as the backbone network and tames it for high predictive accuracy. X-Transformer \cite{2019Taming}, which used the pre-trained model, e.g., BERT \cite{2018BERT}, effectively extracted features from raw texts to improve accuracy significantly. Nevertheless, X-Transformer is not satisfying when it comes to calculating costs. LightXML \cite{2021LightXML} later improved the X-Transformer to make it lighter and faster, and LightXML also reached state-of-the-art. However, LightXML and X-Transformer could further on with a valid fine-tuned strategy. Furthermore, they need to consider the critical label semantics that is easy to find.

The descriptive label semantics need to be utilized more to find the latent space. The sparse linear network is not robust for feature extraction. To this end, we designed a feature extractor using an improved Transformer-based model as the backbone network for the text features. The ability of the Transformer-based model to extract word-level and sentence-level features has been widely recognized. Furthermore, we propose a novel guide network (GUDN) to combine with the feature extractor to improve the performance of XMTC tasks. The guide network can give a precise guide for the text feature and reduce some irrelevant information. Considering a vast semantic gap between text and label, increasing the prediction difficulty, we design a label reinforcement strategy to help out.

The contributions of our work are summarized as follows:
 \begin{itemize}
 	\item A novel guide network (GUDN) that includes two modules of guide and two loss functions is proposed. GUDN instructs the Transformer-based model to capture the label-aware features from the text for exploring the latent space between texts and labels.
 	
 	\item GUDN considers raw label semantics with an artful label reinforcement strategy that is practical and succinct and combines a refined deep pre-trained model to extract features for predictive accuracy.
 \end{itemize}
The rest of the paper is organized as follows: Section 2 examines works related to this study. In Section 3, we provide a detailed description of the proposed methods. We show the experimental results in Section 4. Section 5 summarizes the paper and indicates future work.

\section{Related Work}
Since the inception of the XMTC problem, many approaches have been exquisitely constructed. Generally, there are two methods to deal with the XMTC problem. One is the traditional method which can be further divided into three categories: embedding-based, tree-based, and one-vs-all (OVA) strategies. Another is deep learning methods, which have become popular in recent years. In the following part, we introduce these methods succinctly.

\textbf{Embedding-based methods:}
Embedding-based methods reduce label redundancy under the low-rank assumption to alleviate the storage and computing overhead. We can also think of this behavior as coding, so the training process is naturally a codec process. The low-dimension theory for label embedding was first proposed by \cite{2009}. SLEEC \cite{2015sparse} captures label correlations non-linearly to reduce the adequate number of labels and cluster the data to speed up the training stage. After SLEEC, \cite{2016Robust} improved the optimization strategy and obtained good results. AnnexML \cite{2017AnnexML} also advanced SLEEC, which addressed issues such as unreasonable data partitioning, indirect objective function, and slow prediction speed. \cite{ceiling} proves a new theory that overfitting is the main reason for the poor performance of embedding-based methods. As a result, they develop GLaS to reduce overfitting. However, through embedding and de-embedding, the label information is inevitably lost. In addition, label distribution's ``long-tail'' tends to harm this assumption.

\textbf{Tree-based methods:}
Tree-based methods hierarchically divide label sets to generate a label tree, so if the tree is balanced, the prediction time will be sub-linear, even logarithmic. FastXML is a typical tree-based instance that optimizes the ranking loss function. SwiftXML \cite{swift} adopts the warm-start strategy and uses label features for performance.
\cite{2018A} proposed hierarchical softmax approaches to reduce training time, while Parabel \cite{2018Parabel} introduced and improved probabilistic label trees (PLTs) \cite{liu2013probabilistic}.
CRAFTML \cite{siblini2018craftml} divides the label tree quickly through a modified random forest algorithm. Though a tree's structure reduces the prediction time, accuracy is affected as the tree grows deeper.

\textbf{OVA methods:}
OVA methods such as PPDSparse \cite{yen2017ppdsparse},  Slice \cite{2019Slice}, Dismec and Bonsai \cite{2020Bonsai} are designed to generate a binary classifier for each label, resulting in significantly improved prediction accuracy. PPDSparse via a new loss function to expand the training. Slice's negative sampling technique has become an effective means to deal with the XMTC challenges. Dismec's layer parallelization strategy and Bonsai's label tree structure, to a certain degree, make up for the deficiency of OVA methods that say the model is large and the calculation complex, but that may still be unacceptable in real-world applications.

\textbf{Deep learning methods:}
In consideration of OVA approaches needing too many calculated resources, at the same time, embedding-based methods rely on the low-rank label assumption, and tree-based ways always lead to reduced accuracy and large model size. Furthermore, traditional methods typically input sparse text features, such as BOW features, to the model, which needs to be improved. As mentioned above, we need an advanced methodology to stand around the problem. Deep learning (DL) methods have recently flourished for XMTC tasks. Since the first DL method, XML-CNN \cite{JingzhouLiu2017DeepLF}, was proposed, researchers have devised many DL methods. The current trend suggests that DL is gradually dominating this domain. Our method is based on DL.

APLC-XLNet \cite{2020pretrained} designs a new way for label partitioning called probabilistic label clusters, reducing computation time. \cite{niculescu2017label} gives a label filter accelerated label prediction. \cite{Kalina2016Extreme} raises sparse probability estimates for calculation cost. \cite{jainextreme} improves the loss function, making it more reasonable. \cite{2019data} focuses on the issue of ``long-tail'' distribution, hoping to alleviate this phenomenon and be more accurate.

C2AE first proposed the hypothesis of a latent space between the label and the text and conducted a preliminary exploration. RankAE also thinks text and label features can represent every text, so a common latent space must exist between text and label features. RankAE suggested a margin-based ranking loss and dual-attention mechanism to find the latent space and improve C2AE.
DXML also hopes to find the latent space to help establish the connection between the text and the label. Hence, DXML creatively considers the label structure information and label metadata. They use the raw label when constructing the label graph. LAHA used the attention mechanism with a label co-exist graph to integrate labels and text semantics.

However, the label semantics are drastically reduced due to hiding the semantics by the graph structure in LAHA and DXML. RankAE needs to set a more robust feature extractor for labels rather than a sparse linear neural network. \cite{2021arXiv210800368M,DBLP:journals/corr/abs-2108-00261} emphasized the importance of label metadata such as label structure or label text description. We fully consider the label semantics and use the deep, pre-trained model to extract features directly from the raw labels.

Inspired by the success of deep pre-training models in natural language processing, X-Transformer tamed the pre-trained Transformer-based model to handle XMTC tasks. Considering the X-Transformer's computational complexity and the model's size, LightXML intends to improve it to obtain a light and fast model. LightXML has reached the most advanced level. However, X-Transformer and LightXML needed to explore a better way for fine-tuning. They only rely on the final objective function to fine-tune Transformer-based models, which is still tricky for extreme classification. To figure out the above problems, we design a novel guide network to guide the Transformer-based model further to extract label semantics.

\begin{figure*}[!htb]
\centering
\includegraphics[scale=0.52]{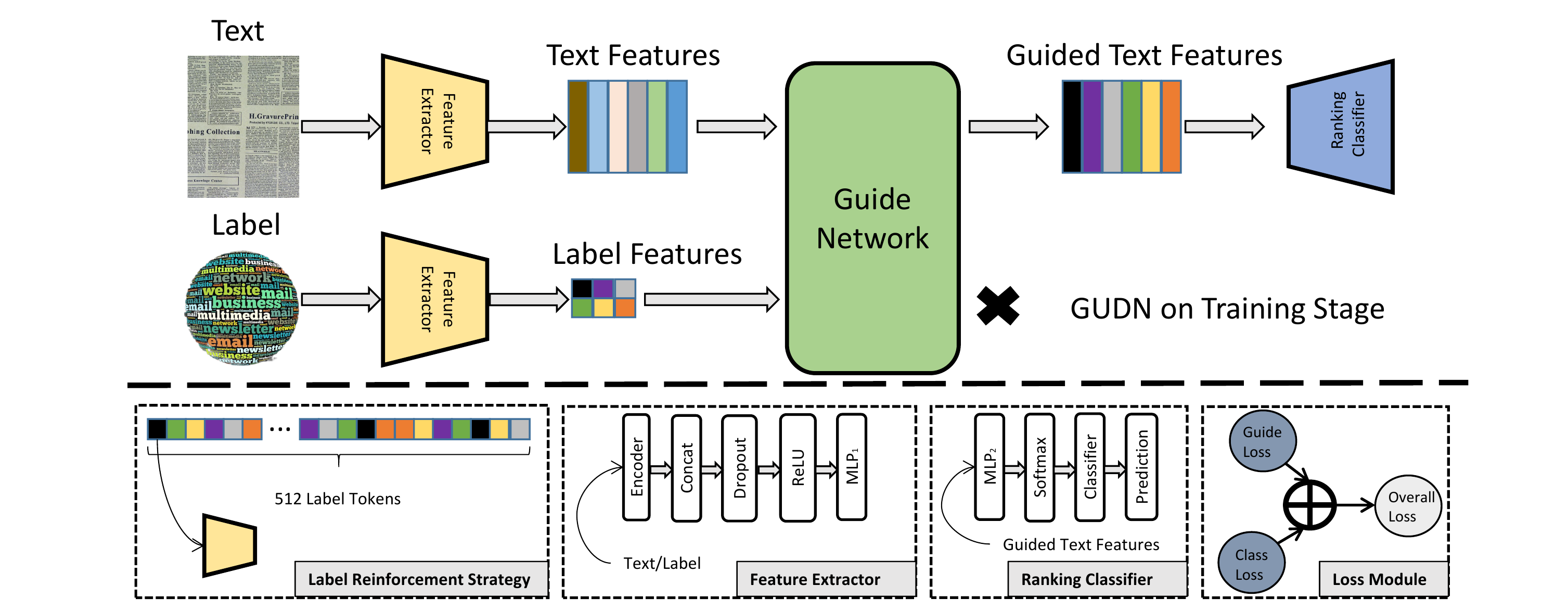}
\caption{GUDN: A novel guide network with label reinforcement for extreme multi-label text classification. The lower part of the figure includes label reinforcement strategy, feature extractor, ranking classifier, and loss module.}
\label{gudn}
\end{figure*}

\section{Proposed Method}
This section gives a detailed description of the proposed method. GUDN is an end-to-end and easily extensible model. The three parts of the proposed model are the feature extractor, guide network, and ranking classifier. Before the extraction process, the label reinforcement strategy is applied to label inputs for better semantic information. In the extraction process, the feature extractor first extracts the features of texts and labels, and then the features of the texts and labels are input into the guide network. A close relationship is established through the guide network for the texts and labels, and the connection is fed back to the feature extractor for continuous optimization. Finally, the ranking classifier obtains accurate semantic information to classify. Figure \ref{gudn} depicts the proposed framework.

\subsection{Preliminaries}
Let $D = \{ (x_1,y_1),(x_2,y_2), ... , (x_n,y_n) \}$ represent the training dataset with $n$ samples where $x_i\in R^d$ is the input of raw text, $y_i\in \{0,1\}^L$ denotes multi-hot vector of true label. The real semantic labels which belong to a sample text are also a part of the input during the training stage. Note that each raw text length is equal to the $d$, and the sum of the number of labels is $L$. We want to find a function $f$ to map $x_i$ and $y_i$. If the $y_{ij}=1$ then the function $f$ will output a high score, wherein
$j\in[ 1 , L ]$. The mapping function $f$ can be expressed as follows:

\begin{equation}
f(x_i,k)= W_k B(x_i),
\label{f}
\end{equation}

where $B(x_i)$ represents the $i$-th text features generated from an encoder $B$ and $W$ is the classifier, usually a fully connected layer. $f$ output the score of $k$-th label. If the score is high, the label is likely to belong to the text.

\subsection{Feature Extractor}
We follow X-Transformer and LightXML using a well-designed BERT to get basic features, as the BERT has proven powerful in natural language processing tasks. The previous methods only take the label multi-hot vector lacking semantic information as the input of the sparse linear network, which is insufficient to find the latent space between labels and texts. To help find the latent space, we use the feature extractor with raw label semantic information.

The structure of the feature extractor shown in Figure \ref{gudn} includes an Encoder layer (BERT), a Concatenation layer, a Dropout layer, a ReLU layer, and an MLP layer. Specifically, an adapted BERT (12 layers and 768 hidden dimensions) is used to extract the original text features. At the same time, the labels share the same BERT with texts to get the label features. Sharing one BERT can significantly reduce the model's size and complexity, accelerating convergence. Text and label features are extracted asynchronously in the training phase but used together to calculate the loss.

Like LightXML, we set a dropout layer with high dropout rates to avoid overfitting. Label description is usually shorter and has less semantic information than text. So, we put together the output of the last eight layers of the ``[CLS]'' token as the extracted features of texts and add two (two is an empirical parameter) more layers of label features to increase semantic information. After the dropout layer, a ReLU activation function and an MLP layer refine the features. The feature extractor $F$ can be defined as follows:

\begin{equation}
E = W_e\sigma(D(f))+b_e ,
\label{eq:fe}
\end{equation}

where $f$ represents the splicing features from the encoder, $D$ is the dropout layer, $\sigma$ is the ReLU. $W_e$ and $b_e$ are the parameters of $MLP_1$.
The outputs of the feature extractor are text features $E_t$ and the label features $E_l$.

\begin{figure}[!htb]
\centering
\includegraphics[scale=0.48]{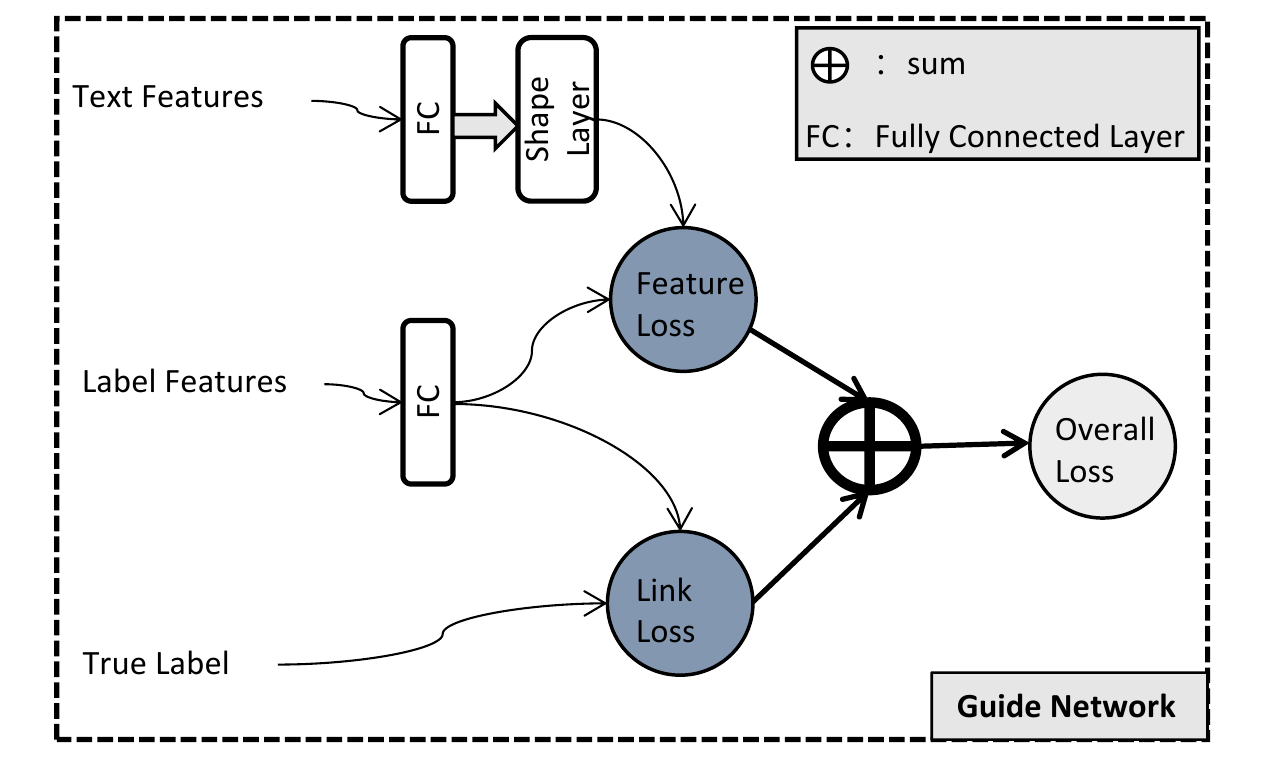}
\caption{The Guide Network.}
\label{gm}
\end{figure}

\subsection{Guide Network}
Relying directly on the simple classified network to link texts to labels is like being lost in the sea without a guide, which is unstable and uncertain. A simple and effective way to solve this problem is to create a guide mechanism for labels and texts. The CLIP \cite{clip} and MICoL \cite{hjw} give a paradigm from the perspective of contrastive learning to deal with the text, which inspires the creation of the guide network. In the GUDN, texts and labels have roles similar to visual and textual input.

In the previous work, C2AE and RankAE provide a bridge between texts and labels, which means they try to find a latent space between texts and labels, and the sparse linear network is trained to guide the classification. However, the bridge is only strong enough with solid label semantics. The guide needs to be disciplined enough with the raw label semantics, and more than using the simple linear network is required to get reasonable label representations. DXML and LAHA consider constructing a label graph structure for prediction, but they ignore the metadata of the label. Furthermore, the graph structure hides the original label semantic in some ways. Consequently, we proposed the guide network to solve the above problems.

Feature extractor is designed to gain semantic-grained features. The detailed module of the guide network is described in Figure \ref{gm}. Guide network has a simple structure. The success of the guide network depends on two guides. The first guide instructs BERT to learn the most representative label features from text features. As a result, the latent space between text and label semantics is discovered effectively. The text and label features from the feature extractor are sent to a fully connected (FC) layer, and the text features still need to pass through a shape layer to match the shape with label features. The above process is the first guide. The other guide can directly establish a mapping relationship between label features and true labels to reduce the pressure on the ranking classifier. For simplicity, we still mark the text features as $E_t$ and the label features as $E_l$.

Finally, the guide network helps us find a certain way from the texts to the actual labels. The loss functions $L_{feature}$ and $L_{link}$ will be the solid bridges in the guide network. Through bridge $L_{feature}$, text space and label space can blend, while labels and label features can be connected through bridge $L_{link}$. The expressions are as follows:

\begin{equation}
L_{feature}(E_t,E_l) = \frac{1}{2n} \sum_{i=1}^n\left \| E_{ti} - E_{li}\right \|^2,
\label{eq:eq1}
\end{equation}

\begin{equation}
L_{link}(y,\hat{y}) =  \sum_{i=1}^n \sum_{j=1}^L -y_{ij}\log (\hat{y_{ij}}) - (1-y_{ij})\log (1-\hat{y_{ij}}).
\label{eq:eq2}
\end{equation}

Equation \ref{eq:eq1} is the mean square error loss (MSE) which is calculated from label features $E_t$ and text features $E_l$, while Equation \ref{eq:eq2} is the binary cross-entropy loss (BCE) calculated from true label $y$ and predicted label $\hat{y}$. $\hat{y}$ is not produced from text features but label features. The sum loss of the guide network $L_{guide}$ is the sum of $L_{feature}$ and $L_{link}$, which is described as follows:

\begin{equation}
L_{guide} =  L_{feature} + L_{link}.
\label{eq:eq3}
\end{equation}

Theoretically, minimizing $L_{guide}$ can make the feature extractor and ranking classifier not dependent on the guide network anymore. We do not use label information during the test stage because the feature extractor and ranking classifier have made it possible to find a way from text to the correct labels alone after being guided by the guide network.

The guide network is not limited to the classification problems in extreme cases described in this paper. It is also suitable for multi-label and multi-category classification, and the network structure is simple and can be easily extended.

\subsection{Ranking Classifier}
The ranking classifier includes an MLP layer, a Softmax layer, and the classifier, as seen in the lower part of Figure \ref{gudn}. The ranking classifier ranks and gets the final result. The formula for the ranking classifier is as follows:

\begin{equation}
y{\prime} = W_c\theta(W_1(E_t)+b_1) ,
\label{eq:rc}
\end{equation}

where $W_1$ and $b_1$ are the parameters of $MLP_2$, $\theta$ represents the softmax layer, and $W_c$ denotes the classifier. For medium-size datasets Eurlex-4K, AmazonCat-13K, and Wiki10-31K, we do not change the original output space of the ranking classifier. However, for the large-scale dataset Wiki-500K, we go after LightXML to adopt a dynamic negative sampling strategy. The $k$ label clusters with the highest recall probability are selected from the output space, and then the candidate labels are selected from clusters.

The final candidate set contains all positive and many ``hard negative'' samples. This strategy not only compresses the output space but also enhances accuracy. In general, label clustering is required according to BOW before dynamic negative sampling, but X-Transformer also gives new insights like positive instance feature aggregation. In this work, we still use BOW for simplicity. The classification loss is also the BCE loss, which can be written as:

\begin{small}
\begin{equation}
L_{class}(y,y{\prime}) =  \sum_{i=1}^n \sum_{j=1}^L -y_{ij}\log (y_{ij}{\prime}) - (1-y_{ij})\log (1-y_{ij}{\prime}),
\label{eq:eq4}
\end{equation}
\end{small}

where $y_i$ is the ground truth, and $y_i{\prime}$ is the labels predicted by text information. They are both $L$-dimensional one-hot vectors.

\subsection{Label Reinforcement Strategy}
We notice a massive length and semantic dichotomy between long text and short-label sequences. In the scenarios of extreme text classification, on the one hand, though the labels can reach more than several thousand, the entire sequences of labels per sample are very short. So, the label input tends to add meaningless, harmful tokens like the ``Padding'' to the end. The long text, however, can reach tens of thousands of characters, causing difficulty in matching.

To this end, we advanced the label input of GUDN. We elaborately set two standards for label combination to make our method progress appreciably. Among the previous methods, \cite{enhancelabel} may be similar to our strategy yet have different aims. \cite{enhancelabel} presented a technique for adding keywords to the input to enhance the label description, but the computing cost for keyword extraction was very high. So we designed a simple approach that can effectively enrich the input of label semantics. In addressing the dichotomy, we use two label reinforcement standards: an ordered label fill method and a disordered label fill method.

For the ordered label fill method, we copy the label sequence of a specific sample. We add the copy sequence to the tail of the original sequence and repeat this operation until the whole sequence's length reaches 512. We adopt a random sample method to complete the input sequence for the disordered method. Specifically, for every label set related to a sample, we shuffle the position of each label and add the set to the tail of the original sequence. This random sampling process follows a normal distribution, so the sequence is finally composed of disordered tokens.

\subsection{Training Process}
GUDN is finally implemented since we have constructed the feature extractor, guide network, and ranking classifier. The objective function $L_overall$, which contains two losses, $L_guide$ and $L_class$, is minimized by GUDN. The sum of two losses caused by the guide network is $L_guide$, and the classification loss is $L_class$. The overall loss function is given in equation (\ref{eq:eq5}).

\begin{equation}
L_{overall} = L_{guide} + L_{class}.
\label{eq:eq5}
\end{equation}

We add the two losses from the guide network to the classification loss because GUDN would be incomplete without any of them. We want the three losses to interact to help GUDN's prediction accuracy achieve the best results. It is hard to optimize when there are all three losses, but GUDN is simple enough, so convergence is possible.

We used label semantics with a label reinforcement strategy during the training stage. We first get text and label features through the feature extractor. They contain the most primitive and crucial semantic information. The text and label features will be entered into the guide network later. Regarding feature loss as a guideline, the guide network helps train the feature extractor and the ranking classifier. Only the feature extractor and the ranking classifier remain after guidance, making the model lighter and suitable for time-sensitive user applications. Finally, GUDN gives the predicted results fast and accurately.

\section{Experiment}
We perform experiments on Linux (Ubuntu 20.04.1). The experiments use four Nvidia GeForce RTX 3090 GPUs with Intel(R) Xeon(R) Gold 6254 CPU @ 3.10GHz to do calculations in parallel, every GPU memory is 24GB, but the training phase occupies less than 20GB. For every experiment, we repeated three times with different random numbers. The results in this paper are the average of these three times experiments.

\begin{table}[!htb]
	\caption{A specific description of datasets. The training and testing set numbers are denoted by $TRN$ and $TST$, respectively. $LBL$ refers to the number of labels. $SPL$ represents the average sample per label, and $LPS$ represents the average label per sample.}
	\centering
    \normalsize
	\setlength{\tabcolsep}{0.88mm}{
	\begin{tabular}{c| c c c c c}
		\hline \hline
		Datasets &$TRN$ &$TST$ &$LBL$ &$SPL$ &$LPS$\\
        \hline
	    Eurlex-4K  &15539   & 3809   & 3993   & 25.73  & 5.31  \\

		AmazonCat-13K &1186239 & 306782 & 13330  & 448.57 & 5.04  \\

		Wiki10-31K &14146   & 6616   & 30938  & 8.52   & 18.64 \\

		Wiki-500K &1813391 & 783743 & 501070 & 23.62  & 4.89 \\
        \hline \hline
	\end{tabular}}
	\label{datasets}
\end{table}

\subsection{Datasets and Evaluation Metrics}
The datasets for the experiments are collected from http://manikvarma.org/downloads/XC/XMLRepository.html \cite{Bhatia16}. Eurlex-4K, AmazonCat-13K, Wiki10-31K, and Wiki-500K were the four representative datasets used. Eurlex-4K is text data about European Union law, containing nearly four thousand labels formed according to EU-ROVOC descriptors. Amazon-13K is a product-to-product recommendation dataset, and labels are the product categories in this dataset. Wiki10-31K and Wiki-500K are excerpts from Wikipedia articles containing about thirty-one thousand and five hundred thousand labels, respectively. Details about the four datasets can be queried in Table \ref{datasets}. It is worth noting that the texts in these datasets are exceedingly long, and generally, their length increases as the label number grows. For example, the article in Wiki-500K is longer than it is in Eurlex-4K.

We use three extensively used metrics in XMTC tasks for comparison. One of these is a simple but intuitive evaluation metric named precision performance at the top ($P@k$). The calculation formula for $P@k$ is as follows:

\begin{equation}
P@k = \frac{1}{k} \sum_{i\in {\rm rank_k}(\hat{y})} y_i ,
\label{eq:eq6}
\end{equation}

where $k$ is constant and it is usually $1$, $3$ or $5$. We rank the predicted results, $haty$, by probability, then select the top $k$ with the highest probability and record their index number. The $P@k$ score is higher if the $k$ indexes have more $1$ values corresponding to the label vector position.

Another metric is the normalized discounted cumulative gain $nDCG@k$, which is defined as follows:

\begin{equation}
DCG@k = \sum_{i\in {\rm rank_k}(\hat{y})} \frac{y_i}{log(i+1)} ,
\label{eq:eq7}
\end{equation}

\begin{equation}
iDCG@k = \sum_{i=1}^{min(k,||y||_0)}\frac{1}{log(i+1)} ,
\label{eq:eq8}
\end{equation}

\begin{equation}
nDCG@k = \frac{DCG@k}{iDCG@k} .
\label{eq:eq9}
\end{equation}

Except for the two metrics mentioned before, the third metric is the propensity-scored performance at the top ($PSP@k$) \cite{jain2016extreme}. This metric allows for avoiding the ``head label strength.'' Typically, in the setting of the XMTC, the number of the ``tail label'' can run high. However, the metrics of $P@k$ and $nDCG@k$ always ignore this phenomenon. So bringing $PSP@k$ to evaluate the model seems more objective. The $PSP@k$ can be defined as follows:

\begin{equation}
PSP@k = \frac{1}{k} \sum_{i\in {\rm rank_k}(\hat{y})} \frac{y_i}{p_i} ,
\label{eq:eq10}
\end{equation}

where $p_i$ is the propensity score for a certain label.

\begin{table*}[!htb]
	\caption{Using $P@k$, we compared the results of experiments with several representative DL methods on Eurlex-4K, AmazonCat-13K, Wiki10-31K, and Wiki-500K. The font in bold indicates the best score, and the underlined font indicates the sub-best score. The `Difference' represents the difference between the experimental results of GUDN and the results of using state-of-the-art methods.}
	\centering
    \normalsize
	\setlength{\tabcolsep}{1.2mm}{
	\begin{tabular}{c c c c c c c c c c}
		\hline
		\hline
		Datasets &P@k &XML-CNN &DXML &AttentionXML &RankAE  &X-Transformer &LightXML &GUDN &Difference\\
		\hline
	      &P@1 &75.32   & -   & 87.12   &  79.52  &  87.22 &\underline{87.63} &\textbf{88.13} &\textbf{+}0.50\\
	     Eurlex-4K &P@3 & 60.14   & -   &  73.99   & 65.14  &  75.12 &\underline{75.89} &\textbf{77.06} &\textbf{+}1.17\\
	      &P@5 &49.21   & -    & 61.92   & 53.18 & 62.90 &\underline{63.36} &\textbf{65.49} &\textbf{+}2.13\\
		\hline
		  &P@1 &93.26 & -  &  95.92  & - & 96.70 &\textbf{96.77} &\underline{96.71} &\textbf{-}0.06\\
		AmazonCat-13K  &P@3 & 77.06 & -  &  82.41  & -  & 83.85 &\underline{84.02} &\textbf{84.19} &\textbf{+}0.17\\
		  &P@5 & 61.40 & -  &  67.31  & -  & \underline{68.58}   &\textbf{68.70}  &67.96 &\textbf{-}0.74\\
		\hline
		 &P@1 & 81.41    & 86.45  &  87.47  & 83.60   & 88.51 &\underline{89.45} &\textbf{89.75}  &\textbf{+}0.30\\
		Wiki10-31K &P@3 & 66.23   &70.88   & 78.48  & 72.07  & \underline{78.71} & \textbf{78.96} &78.58 &\textbf{-}0.38\\
		&P@5 & 56.11   & 61.31   & 69.37  & 62.07   &  69.62 &\underline{69.85} &\textbf{69.86} &\textbf{+}0.01\\
		\hline
		 &P@1 &-  &- &  76.95 & -   & 77.28 &\underline{77.78} &\textbf{77.89} &\textbf{+}0.11\\
		Wiki-500K &P@3 &-  & -  &  58.42 & -   & 57.47 &\underline{58.85} &\textbf{59.15} &\textbf{+}0.30\\
		 &P@5 &-  &- & \textbf{46.14} & -   & 45.31 & 45.57 &\underline{46.01} &\textbf{-}0.13\\
		\hline
		\hline
	\end{tabular}}
	\label{results}
\end{table*}

\begin{table*}[!htb]
	\caption{The experimental results of $nDCG@k$ on Eurlex-4K, AmazonCat-13K, Wiki10-31K, and Wiki-500K compared GUDN with two other significant models. Since the $nDCG@1$ is equal to the $P@1$, we do not list them. The font in bold indicates the best score.}
	\centering
    \normalsize
	\setlength{\tabcolsep}{2.4mm}{
    \begin{tabular}{c |c c| c c |c c |c c}
    		\hline
		\hline
             & \multicolumn{2}{c}{EURLex-4K} & \multicolumn{2}{c}{AmazonCat-13k} & \multicolumn{2}{c}{Wiki10-31K} & \multicolumn{2}{c}{Wiki-500k} \\
             \hline
             & nDCG@3        & nDCG@5        & nDCG@3         & nDCG@5        & nDCG@3          & nDCG@5          & nDCG@3        & nDCG@5        \\ \hline
AttentionXML & 77.44         & 71.53         & 91.17          & 89.48         &   80.61         & 73.79           &    \textbf{76.56}         & \textbf{74.86}         \\ \hline
LightXML     & 78.00         & 71.87         &  91.77        & \textbf{90.58}          & \textbf{81.81}         &  \textbf{74.67}          & 74.71         & 72.19         \\ \hline
GUDN         & \textbf{78.19}         & \textbf{72.75}         &   \textbf{91.98}        &  90.06       & 81.38           & 74.63             & 75.16         & 73.78  \\
    		\hline
		\hline
\end{tabular}}
	\label{results_ndcg}
\end{table*}

\begin{table*}[!htb]
	\caption{The experimental results of $PSP@k$ on Eurlex-4K, AmazonCat-13K, Wiki10-31K, and Wiki-500K compared GUDN with two other significant models. The font in bold indicates the best score.}
	\centering
    \normalsize
	\setlength{\tabcolsep}{3.88mm}{
\begin{tabular}{c c c c c}
		\hline
		\hline
                           & PSP@k & AttentionXML & LightXML & GUDN  \\   \hline
   & PSP@1 & 42.31        & 42.18    & \textbf{43.89} \\
                EURLex-4K                 & PSP@3 & 49.17        & 48.97    & \textbf{50.61} \\
                               & PSP@5 & 52.19        & 53.99    & \textbf{54.98} \\     \hline
 & PSP@1 & 53.76        & \textbf{54.88}    & 54.69 \\
             AmazonCat-13k                  & PSP@3 & 68.72        & 70.21    & \textbf{70.98} \\
                               & PSP@5 & 76.38        & 76.54    & \textbf{77.04} \\
                               \hline
   & PSP@1 & 15.57        & \textbf{16.00}    & 15.99 \\
              Wiki10-31K                  & PSP@3 & 16.80        & \textbf{16.99}    & 16.87 \\
                               & PSP@5 & 17.82        & \textbf{18.97}    & 18.16 \\ \hline
     & PSP@1 & \textbf{34.00}        & 31.99    & 32.71 \\
               Wiki-500k                & PSP@3 & \textbf{44.32}        & 42.00    & 41.91 \\
                               & PSP@5 & \textbf{50.15}        & 46.53    & 46.53   \\		\hline
		\hline
\end{tabular}}
	\label{results_psp}
\end{table*}

\subsection{Experiments Setting}
We set the length of the input texts to 512 due to the limitation of regular BERT. When the labels are short and cannot reach a length of 512, we consider all the labels corresponding to a text as a whole, which can also be thought of as a text, and then feed the entire label sequence to the feature extractor. Using the label reinforcement strategy can increase the label sequence to 512. Note that when the text length is more than 512, we have to decide which part of the text to keep. The reserved portion of the text could be the head, the tail, or even the middle. Some information is lost after dividing, but we only take the first 512 words.

Regarding the training epochs, we set 40 epochs for Eurlex-4K and Wiki10-31K and 20 epochs for the datasets Wiki-500K and AmazonCat-13K due to their large sample numbers. For all the datasets, the training batch size is 8, and the testing batch size is 16.

\subsection{Experimental Results and Discussion}
The experimental results of $P@k$, $nDCG$ and $PSP@k$ are shown in Table \ref{results}, \ref{results_ndcg}, \ref{results_psp} respectively. We experiment with four datasets and compare the results of GUDN with six representative deep-learning approaches. The data of $P@k$ of the six models were obtained from their original papers. As far as possible, we refer to the published experimental data for the results of $nDCG@k$ and $PSP@k$ of AttentionXML \cite{2018AttentionXML} and LightXML, though it is incomplete.

We use the reproduced model test to get results for the data we could not collect. Among these six most representative models, DXML and RankAE were similar to the C2AE, trying to find a latent space between texts and labels, which is also one of the aims of GUDN. Although their achievements have been surpassed, their ideas are still inspiring. Compared with DXML and RankAE, GUDN uses a deep, pre-trained Transformer-based model to directly extract the raw label semantics, which is more conducive to finding the latent space.

To our knowledge, XML-CNN was the first method to use the deep network for the XMTC task. The results of AttentionXML showed that the accuracy was significantly improved compared to XML-CNN. The primary opponents of GUDN are X-Transformer and LightXML, and they are all based on the pre-trained model to encode the text and used to be state-of-the-art. Like X-Transformer and LightXML, the GUDN also uses BERT as the backbone network. With feature loss as the guideline and under the guidance of the guide network with label reinforcement strategy, GUDN also achieves significant results.

After training, all three losses decreased to a considerable extent. Among the three losses, feature loss dropped the most, class loss followed, and link loss was the least obvious. The ablation study also shows how significant the three losses were when learned separately.

One of the core challenges of XMTC includes accuracy. Table \ref{results} shows that the GUDN reaches state-of-the-art on Eurlex-4K, especially for $P@5$. GUDN also gains some advantages on AmazonCat-13K, Wiki-500K, and Wiki10-31K. It is not hard to see that GUDN does not perform as well on AmazonCat-13K, Wiki10-31K, and Wiki-500K as it does on Eurlex-4K. Because of the industry's diverse applications, XMTC methods need to consider training and prediction efficiency. Due to its simple architecture, the GUDN is still competitive regarding model size and calculation time. Limited by the hardware, we cannot reproduce some models, so we cannot get the training time and memory usage of these models for comparison. However, we give the training time and model size of the GUDN. Refer to Table \ref{MT} for specific experimental data.

\begin{figure*}[!htb]
\centering
\includegraphics[scale=0.6]{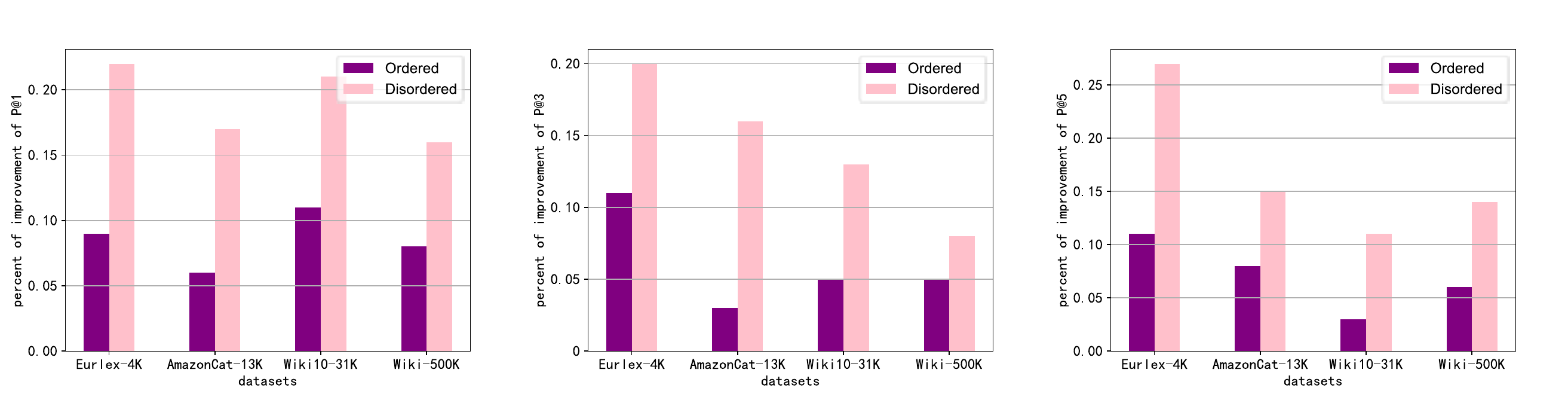}
\caption{On four different datasets, the enhanced effect at $P@1$, $P@3$, and $P@5$ uses the label reinforcement strategy with ordered and disordered settings, respectively.}
\label{labelR1}
\end{figure*}

We conduct experiments on the label reinforcement strategy using GUDN as the fundamental model. The results of two standards of label reinforcement are compared in Figure \ref{labelR1}. These improvements are based on the best performance of GUDN.

The label reinforcement strategy improves matching performance by 0.03\%-0.3\% compared to basic GUDN, whether ordered or disordered. Moreover, under the disordered scenario, the improvements are more pronounced because the label sequence does not have any order in the real world. On the other hand, the Transformer-based model considers the input to have a context relation. So with a disordered label sequence, we can let GUDN learn more features about the label and its semantics.

\begin{table}[!htb]
\caption{This table shows the model size of GUDN in GB and the time required for each epoch of training under the current experimental setting in minutes.}
\setlength{\tabcolsep}{1.2mm}{
\begin{tabular}{ccccc}
\hline
\hline
    &Dataset & Model Size & Training Time \\
    \hline
    &Eurlex-4K  &2.66    &1.50  \\
    \hline
    &AmazonCat-13K  &2.71 &62.16    \\
    \hline
    &  Wiki10-31K     &2.83      &3.47 \\
    \hline
    & Wiki-500K & 3.12     &95.43          \\
    \hline
    \hline
\end{tabular}}
\label{MT}
\end{table}

\begin{figure}[!htb]
\centering
\includegraphics[scale = 0.68]{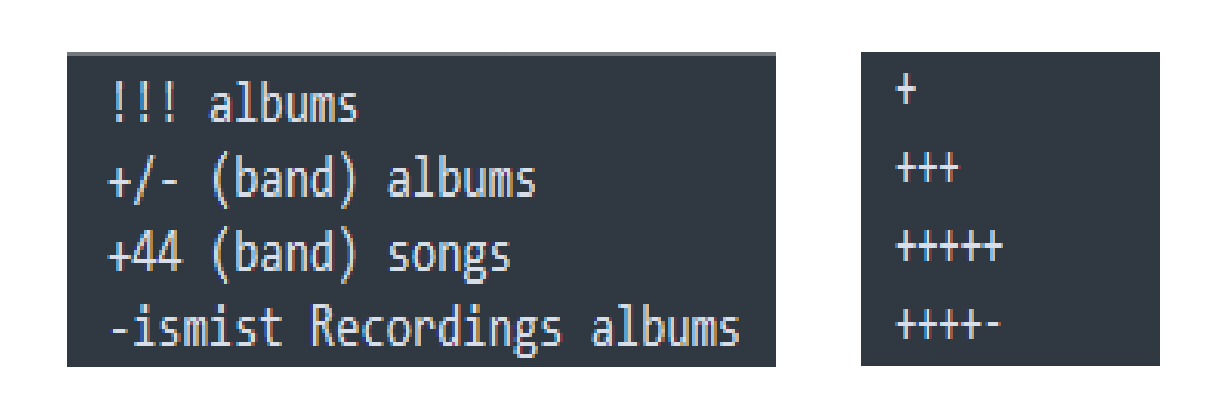}
\caption{Examples of the semantically lacking or confusing labels are taken from Wiki-500K and Wiki10-31K. Each line represents a label.}
\label{lb}
\end{figure}

\subsection{Detail Analysis}
In this section, an in-depth analysis of the experimental data is displayed. We explored why GUDN performs less well than Eurlex-4K on other datasets. According to our research, the label settings in the dataset Eurlex-4K are more consistent with the semantics of the natural language, which has solid semantic features and represents its corresponding text. At the same time, many labels are semantically lacking or confusing in AmazonCat-13K, Wiki10-31K, and Wiki-500K. We call them symbolic labels. Figure \ref{lb} shows examples of symbolic labels, such as '!!! albums' and '++++-'. Some labels consist of low-level semantic characters. Some labels have words with complete semantics but are disturbed by some invalid characters.

Therefore, we conclude that GUDN is sensitive to semantic information, which impairs its performance on datasets with weak label semantics but is favorable for GUDN on datasets with strong label semantics. We further clarify this conclusion in ablation experiments. Moreover, the experiment on the label reinforcement strategy has a similar conclusion.

\begin{table*}[!htb]
\caption{Comparison of ablation results on four datasets for each network module. GUD-F and GUD-L represent the feature guide and link guide, respectively.} \setlength{\tabcolsep}{2.18mm}{
\begin{tabular}{c|ccc|ccc|ccc|ccc}
          \hline \hline
    & \multicolumn{3}{c}{Eurlex-4K} & \multicolumn{3}{c}{AmazonCat-13K} & \multicolumn{3}{c}{Wiki10-31K} & \multicolumn{3}{c}{Wiki-500K} \\
          \hline
    Modules   & P@1      & P@3      & P@5     & P@1       & P@3       & P@5       & P@1      & P@3      & P@5      & P@1      & P@3      & P@5     \\
\hline
BERT      & 86.68    & 75.04    & 63.03   & 96.10     & 82.89     & 67.01     & 88.76    & 77.65    & 68.51    & 77.37    & 58.09    & 45.46   \\
\hline
BERT+GUD-F & \textbf{87.93}   & \textbf{76.64}    & \textbf{65.02}   & \textbf{96.56}     & \textbf{83.98}     & \textbf{67.66 }    & \textbf{89.62}    & \textbf{77.98}   & \textbf{69.67}    & \textbf{77.76}    & \textbf{58.98}    & \textbf{45.69}   \\
\hline
BERT+GUD-L & 87.21    & 75.09    & 63.93   & 96.12     & 83.10     & 67.28     & 89.21    & 77.63    & 68.94    & 77.41    & 58.16    & 45.62 \\ \hline \hline
\end{tabular}}
\label{abl}
\end{table*}

\begin{table}[!htb]
\caption{Experimental results on the Eurlex-4K, AmazonCat-13K, Wiki10-31K, and Wiki-500K datasets, comparing the improvement of $P@k$ over basic BERT using multi-hot vectors and raw labels as input.}
\setlength{\tabcolsep}{1.18mm}{
\begin{tabular}{ccccc}
\hline
\hline
    & & P@k & Raw labels & Multi-hot vectors \\
    \hline
    & & P@1 & +1.25      & +0.03           \\
    &Eurlex-4K & P@3 & +1.60    & +0.32           \\
    & & P@5 & +1.99      & +0.13           \\
    \hline
    &  & P@1 & +0.67      & +0.18           \\
    &  AmazonCat-13K      & P@3 & +0.99      & +0.08           \\
    &         & P@5 & +0.79     & +0.16        \\
    \hline
    &  & P@1 & +0.86      & +0.15           \\
    &  Wiki10-31K       & P@3 & +0.33      & +0.18           \\
    &         & P@5 & +1.16      & +0.10        \\
            \hline
    & & P@1 & +0.76      & +0.19           \\
    &Wiki-500K & P@3 & +0.71   & +0.22           \\
    & & P@5 & +0.29      & +0.10           \\
    \hline
    \hline
\end{tabular}}
\label{4500}
\end{table}

\begin{table*}[!htb]
\caption{A comparison of the effect of different input lengths on accuracy on four datasets.}
\setlength{\tabcolsep}{2.18mm}{
\begin{tabular}{c|ccc|ccc|ccc|ccc}
          \hline \hline
    & \multicolumn{3}{c}{Eurlex-4K} & \multicolumn{3}{c}{AmazonCat-13K} & \multicolumn{3}{c}{Wiki10-31K} & \multicolumn{3}{c}{Wiki-500K} \\
          \hline
   Input length   & P@1      & P@3      & P@5     & P@1       & P@3       & P@5       & P@1      & P@3      & P@5      & P@1      & P@3      & P@5     \\
\hline
512      & 88.01    & \textbf{77.08}    & \textbf{65.37}   & 96.18    & 84.15     & 67.86     & 89.70    & 78.58    & 69.87    & 77.87    & 59.16    & 45.98   \\
\hline
1024 & \textbf{88.12}    & 76.96    & 65.29   & 96.26    & \textbf{84.21}     & \textbf{67.99}     & 89.96    & 78.62    & 69.93    & 77.90    & 59.20    & 46.07   \\
\hline
2048 & 88.03    & 76.90    & 65.13   & 96.25     & 84.07     & 67.79     & 90.08    & 78.71    & 70.03    & 77.94    & 59.29   & 46.12 \\ \hline
4096 & 87.95    & 76.90    & 64.98   & \textbf{96.73}    & 84.06     & 67.62     & \textbf{90.21}    & \textbf{78.83}    & \textbf{70.16}    & \textbf{78.01}    & \textbf{59.31}    & \textbf{46.18} \\ \hline\hline
\end{tabular}}
\label{longP}
\end{table*}

\subsection{Ablation Study}
A single BERT model, BERT with feature guide (helps find the latent space), and BERT with link guide (reduces the pressure of ranking classifier) are tested to prove that the guide network helps BERT effectively extract features to solve the XMTC problem. When learned separately, these experiments also show the importance of the three losses, i.e., feature loss, link loss, and classification loss. Even though there is a relatively slight decrease in link loss, the effect cannot be ignored. It should be mentioned that when we only use a single BERT model, GUDN is equivalent to LightXML, but the fact that we cannot reproduce it with the same accuracy is regrettable. The impact of different modules on accuracy is shown in Table \ref{abl}.

Table \ref{abl} clearly shows that the model employing the guide network outperforms the model using only a single BERT. It also proves that the guide network helps fine-tune BERT to catch the label-aware features in texts and labels to establish a close connection, then find the latent space. According to Table \ref{abl}, the accuracy of the dataset Eurlex-4K increases most obviously, indicating that the guide network is sensitive to label semantics. We also find that both the feature guide and the linked guide contribute to accuracy and are indispensable. The feature guide contributes more to accuracy improvement than the link guide, and the model works best when the two work together.

Table \ref{abl} clearly shows that the model employing the guide network outperforms the model using only a single BERT. It also proves that the guide network helps fine-tune BERT to catch the label-aware features in texts and labels to establish a close connection, then find the latent space. According to Table \ref{abl}, the accuracy of the dataset Eurlex-4K increases most obviously, indicating that the guide network is sensitive to label semantics. We also find that both the feature guide and the linked guide contribute to accuracy and are indispensable. The feature guide contributes more to accuracy improvement than the link guide, and the model works best when the two guides work together.

Besides, it can be found that using raw label semantics on four datasets can converge more quickly than the multi-hot vectors. Because the raw labels are more in line with natural language norms than those simple representations of multi-hot vectors, they provide richer semantic information to explore the latent space between texts and labels.

\subsection{Longformer vs BERT}
We conduct an additional experiment to explore the influence of input text length. The limited input length of most Transformer-based models is an obvious flaw. They have a limit of 512 for the length of the input sequence. The Longformer \cite{longformer} can make the input sequence longer than 512. The length of input of Longformer can reach more than 4096 tokens. We set 512, 1024, 2048, and 4096 as lengths for text input to compare the influence. This operation can be regarded as a text reinforcement strategy. We only choose the header part of texts for simplicity.

Table \ref{longP} shows the influence of the input length. The results may have a little discrepancy because the Longformer has a few subtle structural differences from BERT. For Eurlex-4K and AmazonCat-13K, the predictive precision does not have an expected increase but a slight decrease. The results in Wiki10-31K and Wiki-500K increased appreciably.

The lengths of 512 can cover many texts in Eurlex-4K and AmazonCat-13K, so 512 may be enough for these two datasets. On the other hand, a length of more than 512 with meaningless tokens may have a negative impact. However, the length can be tremendous in Wiki10-31K and Wiki-500K. Thus, the additional input length can give an extra gain in predictive accuracy for an extremely long text, but considering the computing time, it may not be suitable.

\begin{figure*}[!htb]
\centering
\includegraphics[scale=0.48]{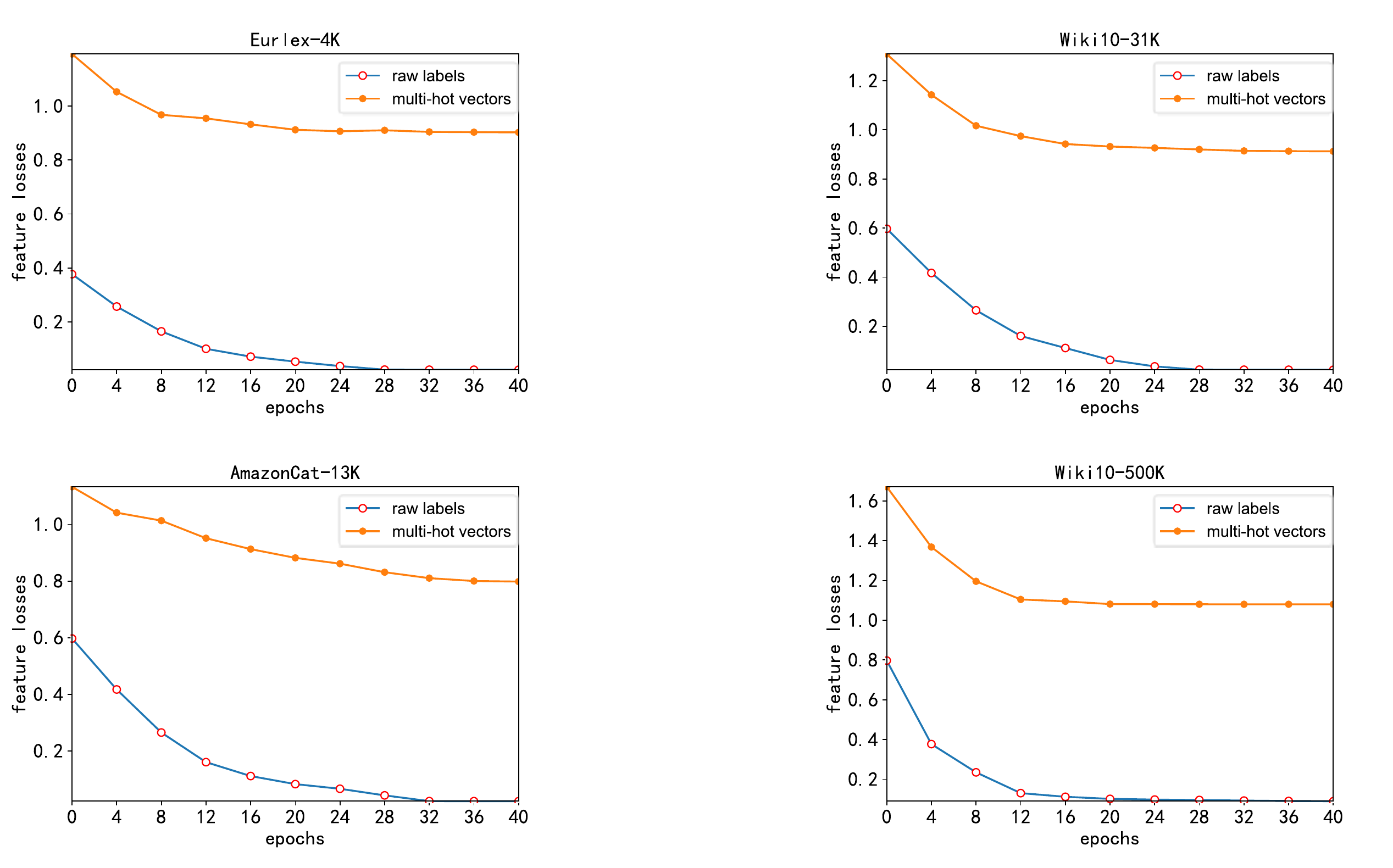}
\caption{An experiment showing the comparison of four datasets on the decay of feature losses with epochs when using raw labels and multi-hot vectors.}
\label{46}
\end{figure*}

\section{Conclusion and Future work}
This paper constructs a novel guide network with a label reinforcement strategy for XMTC tasks. The experimental results show that GUDN achieves competitive performance in multiple datasets, especially in Eurlex-4K. The label reinforcement strategy further improves performance. The ablation experiments prove that the guide network does help fine-tune BERT, and the feature guide and the linked guide play their respective roles. GUDN is sensitive to the semantics of labels, so it has impressive power for datasets with semantically rich labels. Even though GUDN does not perform well on datasets with few semantic labels as it does on datasets with many semantic labels, it still achieved competitive results compared with state-of-the-art results because most real-world labels are composed of natural language.

\cite{DBLP:journals/corr/abs-2103-03494} did an exciting piece of work. They effectively repartitioned the datasets, making the data distribution in the data set more reliable. We might go from there and ask how GUDN fits into their new division rules. Furthermore, we noticed some work on few-shot and short-text, such as \cite{DBLP:conf/kdd/GuptaBPPV21}.

\section*{Acknowledgment}
This work benefited from the support of the Zhejiang Key Laboratory of Intelligent Education Technology and Application. At the same time, this work was supported by the National Natural Science Foundation of China(62077015) and the National Key R\&D Program of China(2022YFC3303600).

\bibliographystyle{cas-model2-names}

\bibliography{GUDN}


\bio{wq}
Qing Wang received a B.E. degree from Zhejiang Wanli University, China, in 2021. Currently, he is a graduate student at the College of Mathematics and Computer Science of Zhejiang Normal University. His research interests include data mining and artificial intelligence.
\endbio
\

\

\bio{zj}
Jia Zhu is a Distinguished Professor at the School of Teacher Education, Zhejiang Normal University, and the Deputy Director of the Key Laboratory of Intelligent Education Technology and Application of Zhejiang Province. He received his Ph.D. degree from the University of Queensland, Australia. His research interests include intelligent education, theoretical algorithms for database and data mining, federated learning, and blockchain with AI.
\endbio

\bio{shj}
Hongji Shu received a B.E. degree from Zhejiang Normal University, China, in 2018. Currently, he is a graduate student at the College of Mathematics and Computer Science of Zhejiang Normal University. His research interests include artificial intelligence and knowledge graph.
\endbio

\

\

\

\

\bio{frank}
Kwame Omono Asamoah received a B.Sc. degree in computer science from the Kwame Nkrumah University of Science and Technology, Ghana, in 2014. He received his master's degree in computer science and technology from the University of Electronic Science and Technology of China in 2018. He had his doctorate degree in computer science and technology from the University of Electronic Science and Technology of China in 2022. He is currently a postdoctoral fellow at Zhejiang Normal University. His current research includes blockchain technology and big data security.
\endbio

\

\

\

\

\

\

\bio{sjy}
Jianyang Shi is a graduate student in the school of teacher education of Zhejiang Normal University. He received his B.E. degree from the Jiangxi Science and Technology Normal University, China. His research interests lie in Classroom multimodal data mining and analysis.
\endbio

\

\

\

\

\bio{zc}
Zhou Cong received a B.E. degree from Zhejiang Wanli University, China, in 2021.  Currently, he is a graduate student at the College of Mathematics and Computer Science of Zhejiang Normal University.   His research interests include data mining and text generation.
\endbio

\end{document}